\documentclass[runningheads,a4paper]{llncs}

\usepackage{amsmath}
\usepackage{amssymb}

\usepackage{graphicx}

\usepackage{cite}
\usepackage{nohyperref}  
\usepackage{url}  

\usepackage{color} 
\usepackage{epsfig}
\usepackage{amsfonts}
\usepackage[linesnumbered,vlined,ruled]{algorithm2e}

\def\squareforqed{\hbox{\rlap{$\sqcap$}$\sqcup$}}
\def\qed{\ifmmode\squareforqed\else{\unskip\nobreak\hfil
\penalty50\hskip1em\null\nobreak\hfil\squareforqed
\parfillskip=0pt\finalhyphendemerits=0\endgraf}\fi}

\newcommand{\keywords}[1]{\par\addvspace\baselineskip
\noindent\keywordname\enspace\ignorespaces#1}

\begin{document}

\mainmatter  


\title{CocoNet: A Deep Neural Network for Mapping Pixel Coordinates to Color Values}

\titlerunning{CocoNet: A Neural Network for Mapping Pixel Coordinates to Color Values}

\author{\vspace{-0.2cm}Paul Andrei Bricman\inst{1}
			\and
        	Radu Tudor Ionescu\inst{2}
}

\institute{\emph{George Co\c{s}buc} National College,
	29-31 Olari, Bucharest, Romania,\\
	\and
    University of Bucharest,
	14 Academiei, Bucharest, Romania\\
	\email{$\{$paubric,raducu.ionescu$\}$@gmail.com}
}

\date{\today}

\toctitle{CocoNet: A Deep Neural Network for Mapping Pixel Coordinates to Color Values}
\tocauthor{Paul Andrei Bricman, Radu Tudor Ionescu}
\maketitle

\begin{abstract}
\vspace{-0,45cm}
In this paper, we propose a deep neural network approach for mapping the 2D pixel coordinates in an image to the corresponding Red-Green-Blue (RGB) color values. The neural network is termed \emph{CocoNet}, i.e. \textbf{co}ordinates-to-\textbf{co}lor \textbf{net}work. During the training process, the neural network learns to encode the input image within its layers. More specifically, the network learns a continuous function that approximates the discrete RGB values sampled over the discrete 2D pixel locations. At test time, given a 2D pixel coordinate, the neural network will output the approximate RGB values of the corresponding pixel. By considering every 2D pixel location, the network can actually reconstruct the entire learned image. It is important to note that we have to train an individual neural network for each input image, i.e. one network encodes a single image only. To the best of our knowledge, we are the first to propose a neural approach for encoding images individually, by learning a mapping from the 2D pixel coordinate space to the RGB color space. Our neural image encoding approach has various low-level image processing applications ranging from image encoding, image compression and image denoising to image resampling and image completion. We conduct experiments that include both quantitative and qualitative results, demonstrating the utility of our approach and its superiority over standard baselines, e.g. bilateral filtering or bicubic interpolation. Our code is available at https://github.com/paubric/python-fuse-coconet.
\keywords{neural networks, image encoding, image denoising, image resampling, image restoration, image completion, image inpainting}
\end{abstract}

\section{Introduction}
\vspace{-0,3cm}

After the success of the AlexNet~\cite{Hinton-NIPS-2012} convolutional neural network (CNN) in the ImageNet Large Scale Visual Recognition Challenge (ILSVRC)~\cite{Russakovsky2015}, deep neural networks~\cite{Simonyan-ICLR-14,Szegedy-CVPR-2015,He-CVPR-2016} have been widely adopted in the computer vision community to solve various tasks ranging from object detection~\cite{Liu-ECCV-2016,Girshick-ICCV-2015,Ren-NIPS-2015,Redmon-CVPR-2016,Soviany-SYNASC-2018} and recognition~\cite{Simonyan-ICLR-14,Ren-NIPS-2015,He-CVPR-2016} to face recognition~\cite{Parkhi-BMVC-2015} and even image difficulty estimation~\cite{Ionescu-CVPR-2016}. In the usual paradigm, a deep neural network (DNN)~\cite{Goodfellow-MITPress-2016} is trained on thousands of labeled images in order to learn a function that classifies the input images into multiple classes, according the labels associated to the images. In this paper, we propose a different paradigm in which the neural network is trained on a single image in order to memorize (encode) the image by learning a function that maps the 2D pixel coordinates to the corresponding values in the Red-Green-Blue (RGB) color space.  At test time, given a 2D pixel coordinate, the neural network will output the approximate RGB values of the corresponding pixel. By considering every 2D pixel location, the network can actually reconstruct the entire learned image at the original scale. By taking pixel locations at different intervals, the network can also reconstruct the image at various scales. We coin the term \emph{CocoNet} for our neural network, i.e. \textbf{co}ordinates-to-\textbf{co}lor \textbf{net}work. To the best of our knowledge, we are the first to propose a neural approach for encoding images individually, by learning a mapping from the 2D pixel coordinate space to the RGB color space. Our novel paradigm enables the application of neural networks for various low-level image processing tasks, without requiring a training set of images. Applications include, but are not limited to, image encoding, image compression, image denoising, image resampling, image restoration and image completion. In this paper, we focus on only three of these tasks, namely image denoising, image upsampling (super-resolution) and image completion. First of all, we present empirical evidence that CocoNet is able to surpass standard image denoising methods on the CIFAR-10 data set~\cite{Krizhevsky-2009}. 
Second of all, we show that our approach can obtain better upsampling results compared to bicubic interpolation on a set of images usually employed in image super-resolution, known as Set5~\cite{Bevilacqua-BMVC-2012}. We also present qualitative results for the task of image completion on Set5.

The rest of this paper is organized as follows. We present related works in Section~\ref{sec_RelatedWork}. We describe our approach in Section~\ref{sec_Method}. We present the results of our image denoising, image upsampling and image completion results in Section~\ref{sec_Experiments}. Finally, we draw our conclusion and discuss future work in Section~\ref{sec_Conclusion}.

\vspace{-0,3cm}
\section{Related Work}
\label{sec_RelatedWork}
\vspace{-0,2cm}

To the best of our knowledge, there are no previous scientific works that propose to learn a mapping of the pixel coordinates to the corresponding pixel color values using neural networks. However, there are numerous neural models that learn a mapping from image pixels to a set of classes~\cite{Hinton-NIPS-2012,Simonyan-ICLR-14,Szegedy-CVPR-2015,He-CVPR-2016} or from pixels to pixels~\cite{Baig-NIPS-2017,Balle-ICLR-2017,Cavigelli-IJCNN-2017,Dong-ECCV-2014,Dumas-ICASSP-2017,Kim-CVPR-2016,Shi-CVPR-2016,Ledig-CVPR-2016,Mao-NIPS-2016,Wu-ACCV-2016,Sajjadi-ICCV-2017,Yamanaka-ICONIP-2017,Minnen-ICIP-2017,Prakash-DCC-2017,Toderici-CVPR-2017,Zhao-MTA-2017,Iizuka-TOG-2017,Yang-CVPR-2017}. The neural models that map pixels to pixels are usually applied on tasks such as image compression~\cite{Baig-NIPS-2017,Balle-ICLR-2017,Cavigelli-IJCNN-2017,Dumas-ICASSP-2017,Minnen-ICIP-2017,Prakash-DCC-2017,Toderici-CVPR-2017}, image denoising and restoration~\cite{Mao-NIPS-2016,Wu-ACCV-2016,Zhao-MTA-2017}, image super-resolution~\cite{Dong-ECCV-2014,Kim-CVPR-2016,Shi-CVPR-2016,Ledig-CVPR-2016,Mao-NIPS-2016,Wu-ACCV-2016,Sajjadi-ICCV-2017,Yamanaka-ICONIP-2017}, image completion~\cite{Iizuka-TOG-2017,Yang-CVPR-2017} and image generation~\cite{Oord-ICML-2016,Huang-CVPR-2017}. Since our approach, CocoNet, can be applied to similar tasks, we consider that the neural models that map pixels to pixels are more closely related to ours. It is important to mention that, different from all these state-of-the-art models, CocoNet does not require a training set of images, as it operates on a single image at once. This represents both an advantage and a disadvantage of our approach. CocoNet can be applied to a given task even if a training set is not available, but models trained a set of images can learn important patterns that generalize well from one image to another. Since CocoNet does not use a set of training images (it just learns a continuous representation of the test image it is applied on), we believe that it is unfair to compare it with supervised models that benefit from the access to a set of training images. We next present related (but supervised) models for each individual task.

\noindent
{\bf Image denoising and restoration.}
Image restoration is concerned with the reconstruction of an original (uncorrupted) image from a corrupted or incomplete one. Typical corruptions include noise, blur and downsampling. Hence, image denoising and image super-resolution can be regarded as sub-tasks of image restoration. There are models that address multiple aspects of image restoration~\cite{Mao-NIPS-2016,Wu-ACCV-2016,Zhao-MTA-2017}, but there are some models that deal with specific sub-tasks, e.g. image super-resolution~\cite{Dong-ECCV-2014,Kim-CVPR-2016,Shi-CVPR-2016,Ledig-CVPR-2016,Mao-NIPS-2016,Wu-ACCV-2016,Sajjadi-ICCV-2017,Tang-IJMLC-2011,Yamanaka-ICONIP-2017}. Zhao et al.~\cite{Zhao-MTA-2017} propose a deep cascade of neural networks to solve the inpainting, deblurring, denoising problems at the same time. Their model contains two networks, an inpainting generative adversarial network (GAN) and a deblurring-denoising network. Wu et al.~\cite{Wu-ACCV-2016} propose a novel 3D convolutional fusion method to address both image denoising and single image super-resolution. To address the same tasks, Mao et al.~\cite{Mao-NIPS-2016} describe a fully convolutional encoding-decoding framework composed of multiple layers of convolution and deconvolution operators.

\noindent
{\bf Image super-resolution.}
Super-resolution techniques reconstruct a higher-resolution image from a low-resolution image. Tang et al.~\cite{Tang-IJMLC-2011} employ local learning and kernel ridge regression to enhance the super-resolution performance of nearest neighbor algorithms. Dong et al.~\cite{Dong-ECCV-2014} propose a CNN model that learns an end-to-end mapping between the low and high-resolution images. Kim et al.~\cite{Kim-CVPR-2016} propose a deeply-recursive convolutional network with skip connections and recursive-supervision for image super-resolution. Ledig et al.~\cite{Ledig-CVPR-2016} present a GAN model for image super-resolution capable of inferring photo-realistic natural images for $4\times4$ upscaling factors. Sajjadi et al.~\cite{Sajjadi-ICCV-2017} employ fully-convolutional neural networks (FCN) in an adversarial training setting. Yamanaka et al.~\cite{Yamanaka-ICONIP-2017} propose a highly efficient and faster CNN model for single image super-resolution by using skip connection layers and network in network modules.

\noindent
{\bf Image compression.}
Image compression is applied to digital images in order to reduce the cost for storing and transmitting the images. Lossy image compression methods usually obtain better compression rates, but they introduce compression artifacts, which are not desirable in some cases, e.g. medical images. Baig et al.~\cite{Baig-NIPS-2017} study the design of deep architectures for lossy image compression. Interestingly, they show that learning to inpaint (from neighboring image pixels) before performing compression reduces the amount of information that must be stored. Ball{\'e} et al.~\cite{Balle-ICLR-2017} describe an image compression method based on three sequential transformations (nonlinear analysis, uniform quantization and nonlinear synthesis), all implemented using convolutional linear filters and nonlinear activation functions. Cavigelli et al.~\cite{Cavigelli-IJCNN-2017} present a deep CNN with hierarchical skip connections and a multi-scale loss function for artifact suppression during image compression, while Minnen et al.~\cite{Minnen-ICIP-2017} combine deep neural networks with quality-sensitive bit rate adaptation using a tiled network. Both Cavigelli et al.~\cite{Cavigelli-IJCNN-2017} and Minnen et al.~\cite{Minnen-ICIP-2017} report image compression results that are better than the JPEG standard. Dumas et al.~\cite{Dumas-ICASSP-2017} address image compression using sparse representations, by proposing a stochastic winner-takes-all auto-encoder in which image patches compete with one another when their sparse representation is computed. Prakash et al.~\cite{Prakash-DCC-2017} design a technique that makes JPEG content-aware by training a deep CNN model to generate a map that highlights semantically-salient regions so that they can be encoded at higher quality as compared to background regions. Toderici et al.~\cite{Toderici-CVPR-2017} present several recurrent neural network (RNN) architectures that provide variable compression rates during deployment without requiring retraining. Their approach is able to outperform JPEG across most bit rates.

\noindent
{\bf Image completion.}
Image completion or image inpainting refers to the task of filling in missing or corrupted parts of images. Iizuka et al.~\cite{Iizuka-TOG-2017} employ a FCN to complete images of various resolutions by filling in missing regions of arbitrary shape. To train their model, the authors use global and local context discriminators that distinguish real images from completed ones. Yang et al.~\cite{Yang-CVPR-2017} propose a multi-scale neural patch synthesis approach based on joint optimization of image content and texture constraints. The approach of Yang et al.~\cite{Yang-CVPR-2017} has two important advantages, namely it preserves contextual structures and produces high-frequency details.

\noindent
{\bf Image generation.}
Perhaps more difficult than image completion is the task of generating entire images that look natural to the human eye. Pixel Recurrent Neural Networks~\cite{Oord-ICML-2016} sequentially predict the conditional distribution over the possible pixel values in an image along the two spatial dimensions, given the scanned context around each pixel. The Pixel RNN model consists of up to twelve fast two-dimensional Long Short-Term Memory (LSTM) layers. Huang et al.~\cite{Huang-CVPR-2017} propose a novel generative model named Stacked Generative Adversarial Networks (SGAN), which is trained to invert the hierarchical representations of a bottom-up discriminative network. They first train each model in the stack independently, and then train the whole model end-to-end.

\vspace{-0,2cm}
\section{Method}
\label{sec_Method}
\vspace{-0,2cm}

\subsection{Intuition}
\vspace{-0,2cm}

The human brain is able to solve an entire variety of tasks ranging from object detection to language understanding, by learning to detect, recognize and understand complex patterns in the input data provided by senses such as sight, hearing or touch. Since the brain has an impressive capacity in solving such complex tasks, researchers have tried to mimic the brain capacity through neural networks~\cite{nn-tricks-2012}, which are a mathematical model of the brain. Most neural network models~\cite{lecun-bottou-ieee-1998,Hinton-NIPS-2012,He-CVPR-2016} are designed to solve classification tasks, but, as the human brain, neural networks also have the capacity to memorize data. For classification problems, memorizing the training data (overfitting) is not desired, as it affects the generalization capacity of the neural model. On the other hand, memory is an important function of the brain and it is useful in solving various tasks, for example in remembering (mapping) important locations in the world, e.g. your home or the nearby grocery store. Although neural networks with memorization capacity have been designed before~\cite{Sukhbaatar-NIPS-2015,Graves-ARXIV-2014} to solve tasks such as question answering and language modeling, these models are based on an external memory tape. Since our main goal is to memorize the provided input data, we exploit the intrinsic memorization capacity of the neural networks, in a different manner than all previous works. As the human brain is able to remember various details corresponding to important locations in the world, we design a neural network that is able to remember the color values corresponding to pixel coordinates in an image.

\vspace*{-0.2cm}
\subsection{CocoNet}
\vspace*{-0.1cm}

\begin{figure}[!t]
\begin{center}
\includegraphics[width=0.75\columnwidth]{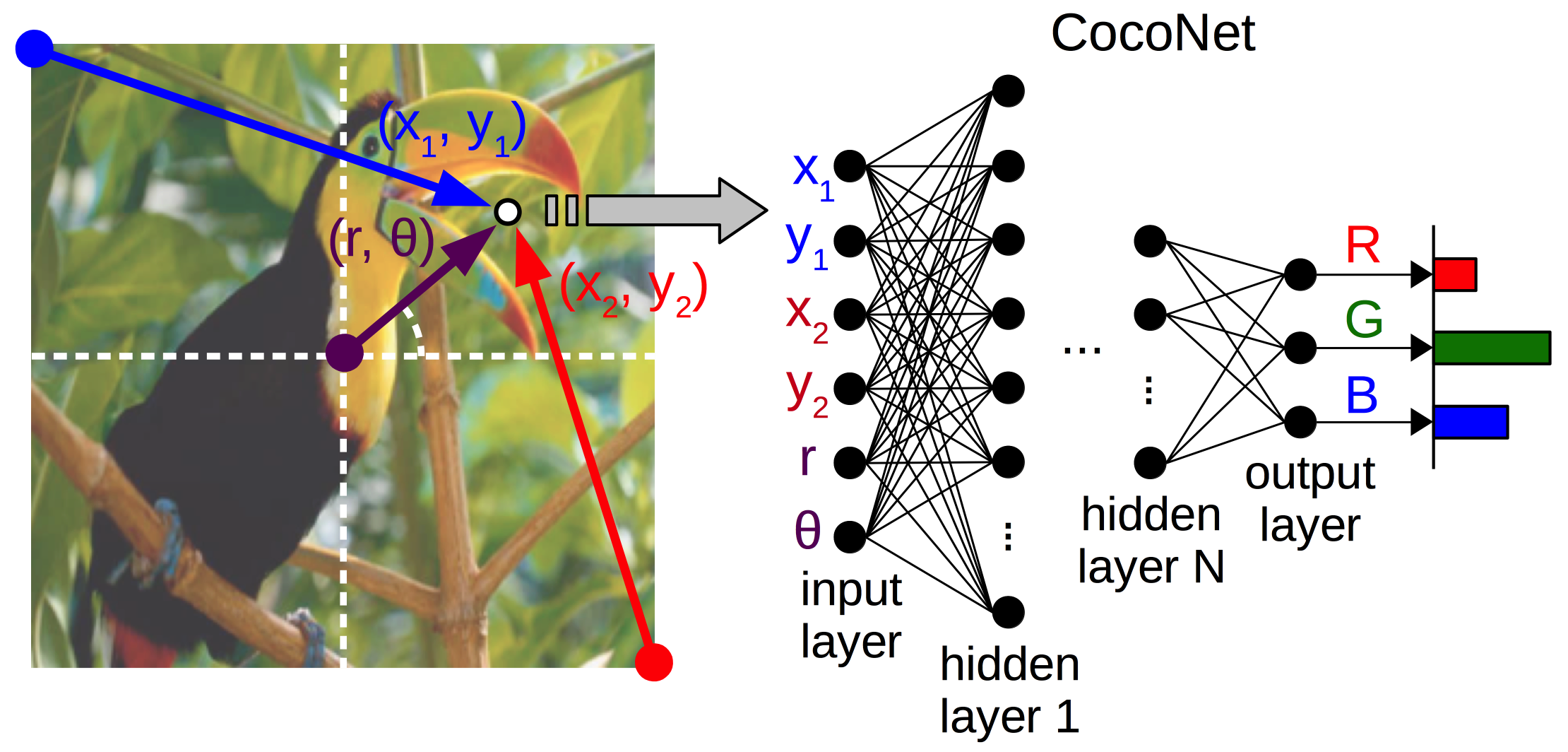}
\end{center}
\vspace*{-0.5cm}
\caption{The generic architecture of CocoNet. The input is given by the pixel coordinates represented in three different coordinate systems (two Cartesian systems and one polar system). The depth of the network can vary, depending on the application. Upon training, the output layer reproduces (with some degree of approximation) the RGB color values for  the pixel coordinates provided as input. Best viewed in color.}
\label{fig_coconet_architecture}
\vspace*{-0.3cm}
\end{figure}

We propose a deep neural network approach that learns a mapping function $f$ from the pixel coordinates in an image to the corresponding RGB color values. The input layer of our neural network is actually composed of $6$ neurons which receive 6D pixel coordinate-related information in the form of a pair of polar coordinates $(r, \theta)$ with the origin in the center of the image and two pairs of Cartesian coordinates $(x_1, y_1)$ and $(x_2, y_2)$. One pair of Cartesian coordinates has the origin in a corner (bottom right) of the image that is diagonally opposite from the corner (top left) taken as origin for the other pair of Cartesian coordinates. It is clear that, from a single pair of coordinates, we can derive every other pair of coordinates through simple transformations. Although the original input is just 2D, i.e. $(x_1, y_1)$, we observed that CocoNet learns a better mapping if we provide the pixel coordinates using multiple coordinate systems. For this reason, we choose to describe each pixel in three different coordinate systems, as shown in Figure~\ref{fig_coconet_architecture}. The input and the output values (targets) are normalized to the $[0, 1]$ interval, for a more efficient optimization during training. The mapping function $f$ learned by our neural network is defined as follows:
\begin{equation}\label{eq_mapping}
f: \mathbb{R}^6 \longrightarrow [0, 1] \times [0, 1] \times [0, 1].
\end{equation}

The generic architecture of our deep neural network consists of fully-connected layers throughout the entire model. Depending on the application, the hidden layers can have a different architecture in terms of depth and width. For image compression, we need to use a thinner architecture with a reduced number of hidden layers. We can employ deeper and wider architectures for image denoising, resampling and completion. The output layer is identical in all applications, consisting of $3$ neurons that output the RGB color values corresponding to the input pixel coordinates, as illustrated in Figure~\ref{fig_coconet_architecture}. Hidden layers are based on the hyperbolic tangent (tanh) activation function, while the final layer is based on the sigmoid transfer function, since the output needs to be in the $[0,1]$ interval.

Each training sample is formed of a 6D input feature vector (composed of pixel coordinates expressed in three different coordinate systems) and a 3D target vector (composed of RGB color values). In this formulation, our neural network can learn to map pixel coordinates to color values for a single image. Each pixel in the image can be taken as training sample, and, for most applications, we consider all pixels as training samples. However, for image completion, we intentionally leave out a region (patch) of the input image that needs to be later reconstructed by the network (at test time). We train the neural network using learning rates between $10^{-4}$ and $10^{-5}$, which provide a more stable convergence during training. We use the mean squared error (MSE) loss function and we optimize the parameters using the Adam stochastic gradient descent algorithm. During training, CocoNet inherently learns a continuous function $f$ from a discrete input image (represented through pixels). Since the learned mapping is continuous, CocoNet can naturally reproduce the image at any scale, at test time. Therefore, our neural network model can be used for image resampling, without requiring any additional developments. Another important aspect is that the neural network implicitly learns a mapping function $f$ that is smooth, hence it will automatically reduce noisy pixel values. Therefore, CocoNet can also be used for image denoising, without any additional changes.

\begin{figure}[!t]
\begin{center}
\includegraphics[width=0.88\columnwidth]{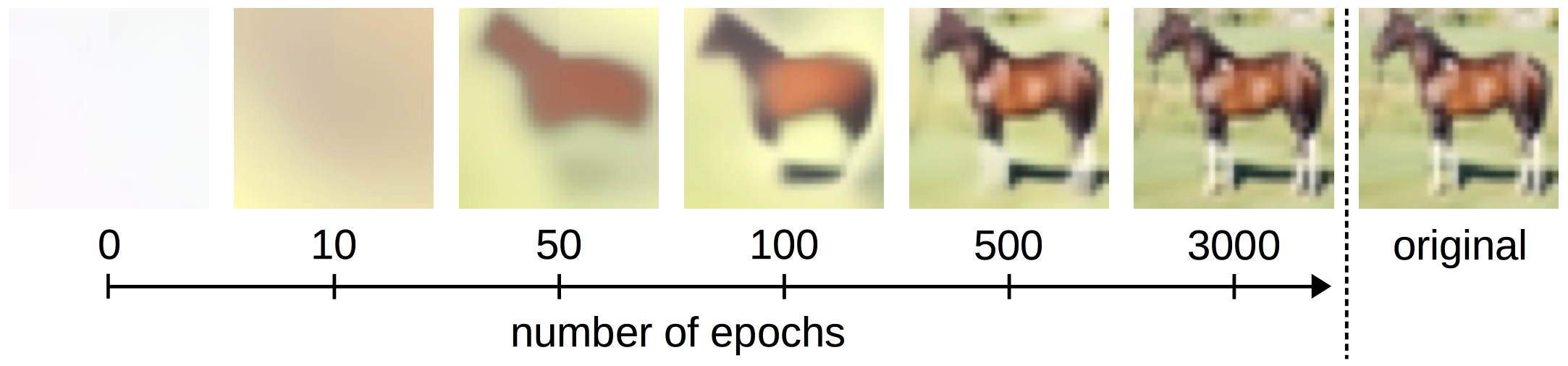}
\end{center}
\vspace*{-0.45cm}
\caption{An image from the CIFAR-10 data set~\cite{Krizhevsky-2009} reconstructed (memorized) by CocoNet, at various stages during the training process. After $3000$ epochs, the memorized image is nearly identical to the original (input) image. Best viewed in color.}
\label{fig_coconet_training}
\vspace*{-0.4cm}
\end{figure}

For a better understanding of how CocoNet learns to memorize the input image within its own layers and weights, we provide a memorized image at various intermediate stages of the training process, in Figure~\ref{fig_coconet_training}. After initializing the network with random weights (close to zero), we can observe that it produces an image that is almost entirely white. After $10$ epochs, the network learns to approximate the average color of the input image. After $100$ epochs, we can already distinguish the main object in the image (a horse). After completing the training process, the image reproduced by the network is very close to the original image. There are very small differences that are indistinguishable to the naked eye. We can observe that during the training process, CocoNet starts by memorizing  a smooth version of the input image, which is gradually refined with increasingly finer details. This comes to support our statement that the neural network tends to learn a continuous and smooth mapping function $f$.

\vspace{-0,2cm}
\section{Experiments}
\label{sec_Experiments}
\vspace{-0,2cm}

\begin{table}[!tpb]
\footnotesize{
\begin{center}
\caption{\label{Tab_Denoising_Results} The PSNR and SSIM metrics of various image denoising filters (mean, Gaussian, median, bilateral) versus a CocoNet architecture with 15 layers. The reported PSNR and SSIM values are averages on the CIFAR-10 test set~\cite{Krizhevsky-2009}. The results are obtained for two Gaussian noise levels with variance $10$ and $20$, respectively. The best result for each metric is highlighted in bold.}
\begin{tabular}{|l|c|c|c|c|}
\hline
\bf Method 							& \multicolumn{2}{|c|}{\bf Noise variance $\mathbf{=10}$} 	& \multicolumn{2}{|c|}{\bf Noise variance $\mathbf{=20}$}\\
\cline{2-5}
                                    & \bf \;\;\;PSNR\;\;\;  & \bf SSIM                  & \bf \;\;\;PSNR\;\;\;      & \bf SSIM\\
\hline
Noisy image 			            & $28.26$				& $0.8928$                  & $22.36$					& $0.7437$\\
$3 \times 3$ mean filter            & $25.47$				& $0.8598$                  & $24.56$					& $0.8216$\\
$5 \times 5$ mean filter            & $21.80$				& $0.6768$                  & $21.60$					& $0.6629$\\
$3 \times 3$ Gaussian filter        & $27.33$				& $0.9057$                  & $25.83$					& $0.8583$\\
$5 \times 5$ Gaussian filter        & $24.96$				& $0.8436$                  & $24.35$					& $0.8160$\\
$3 \times 3$ median filter          & $26.51$				& $0.8732$                  & $24.72$					& $0.8120$\\
$5 \times 5$ median filter          & $22.97$				& $0.7269$                  & $22.43$					& $0.6943$\\
$3 \times 3$ bilateral filter       & $31.24$				& $0.9475$                  & $26.46$					& $0.8629$\\
$5 \times 5$ bilateral filter       & $30.60$				& $0.9413$                  & $\mathbf{27.49}$			& $0.8902$\\
CocoNet (15 layers)                 & $\mathbf{31.25}$		& $\mathbf{0.9503}$         & $27.22$					& $\mathbf{0.8925}$\\
\hline
\end{tabular}
\end{center}
}
\vspace{-0,6cm}
\end{table}

\subsection{Data Sets}
\vspace{-0,1cm}

We use the CIFAR-10 data set~\cite{Krizhevsky-2009} in the image denoising experiments. The CIFAR-10 data set consists of $60$ thousand $32\times32$ color images which are divided into $10$ mutually exclusive classes, with $6000$ images per class. There are $50$ thousand training images and $10$ thousand test images. 

We perform image resampling and completion experiments on Set5~\cite{Bevilacqua-BMVC-2012}, a commonly used data set for image super-resolution~\cite{Kim-CVPR-2016,Ledig-CVPR-2016,Yamanaka-ICONIP-2017}. The Set5 data set consists of $5$ images with various sizes between $228$ and $512$ pixels.

\vspace{-0,2cm}
\subsection{Evaluation}
\vspace{-0,1cm}

The image denoising and resampling results are evaluated against the original images, in terms of the peak signal-to-noise ratio (PSNR) and the structural similarity index (SSIM). The PSNR is the ratio between the maximum possible power of a signal and the power of corrupting noise that affects the fidelity of its representation. Many researchers~\cite{Wang-TIP-2004,Wang-SPM-2009} argue that PSNR is not highly indicative of the perceived similarity. The SSIM metric~\cite{Wang-TIP-2004} aims to address this shortcoming by taking texture into account.



\vspace{-0,2cm}
\subsection{Image Denoising Results}
\vspace{-0,1cm}

\begin{figure}[!t]
\begin{center}
\includegraphics[width=0.95\columnwidth]{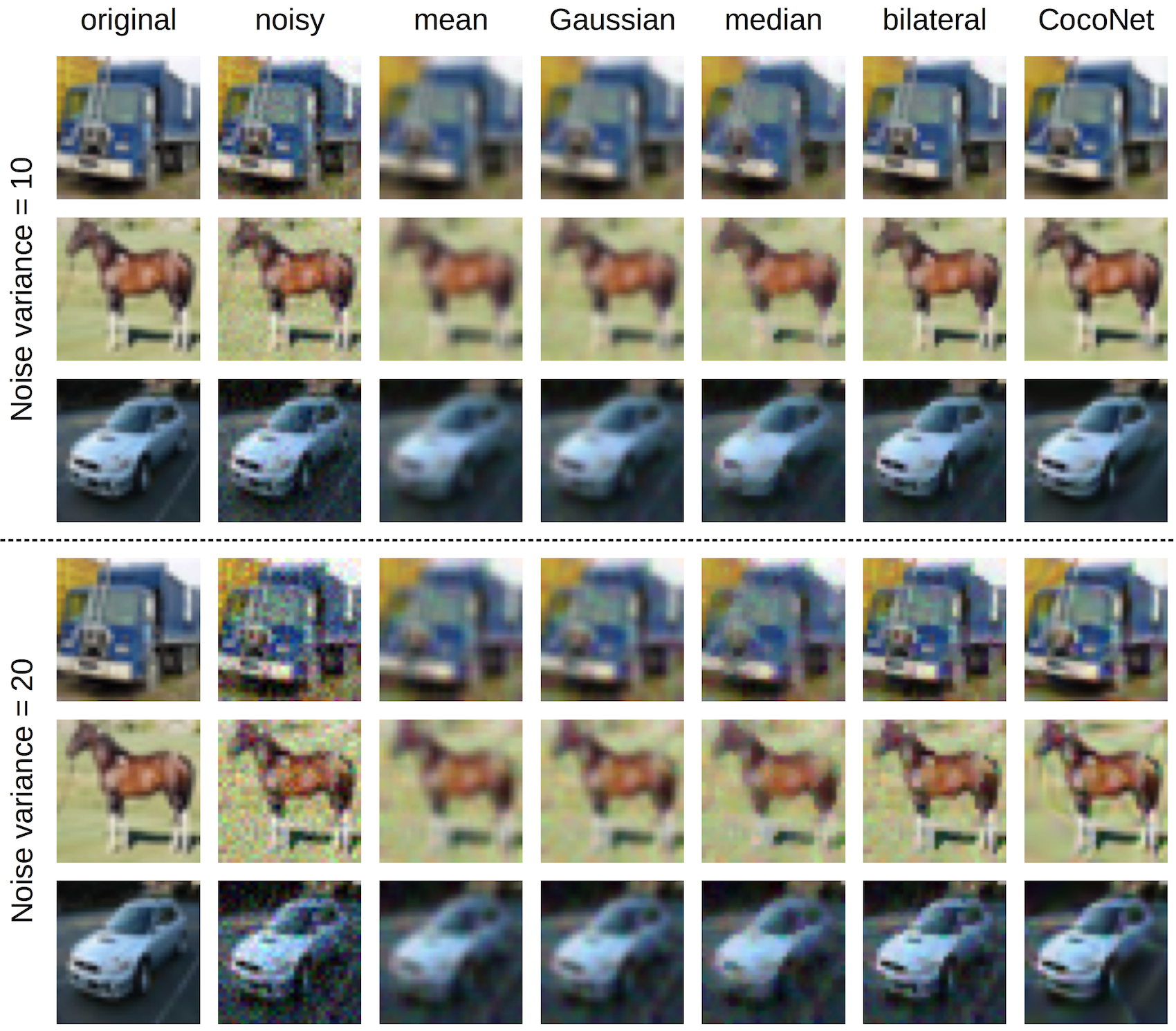}
\end{center}
\vspace*{-0.45cm}
\caption{Three original, noisy and denoised images selected from the CIFAR-10 data set~\cite{Krizhevsky-2009}. The output images resulted after applying various filters (mean, Gaussian, median, bilateral) are compared with the output images produced by a CocoNet architecture with 15 layers. Best viewed in color.}
\label{fig_denoising}
\vspace*{-0.2cm}
\end{figure}

For image denoising, we consider several filters (mean filter, Gaussian, median and bilateral) as baseline denoising methods. For each filter, we consider filter sizes of $3\times3$ and $5\times5$. The baselines methods are compared with a CocoNet architecture with 15 layers, each of $200$ neurons. To generate input images, we apply Gaussian noise over the CIFAR-10 test images, using two noise levels with variance $10$ and $20$, respectively. Table~\ref{Tab_Denoising_Results} presents the results of the considered baseline filters versus CocoNet. The $3\times3$ filters perform better for images with noise variance $10$, while the $5\times5$ filters perform better for images with noise variance $20$. In terms of the SSIM metric, CocoNet surpasses all filters for both noise variances. Although the bilateral filter obtains a slightly better PSNR value for noisy images with variance $20$, it seems that CocoNet is better correlated with the similarity perceived by humans, according to the SSIM metric. Figure~\ref{fig_denoising} shows that the output images of CocoNet are indeed more similar to the original CIFAR-10 images than all filters.

\vspace{-0,2cm}
\subsection{Image Resampling Results}
\vspace{-0,1cm}

\begin{table}[!t]
\footnotesize{
\begin{center}
\caption{\label{Tab_Resampling_Results} The PSNR and SSIM metrics for image upsampling on Set5~\cite{Bevilacqua-BMVC-2012} using two baseline methods (bicubic interpolation, Tang et al.~\cite{Tang-IJMLC-2011}) versus a CocoNet architecture with 10 layers. The upsampling scale factor is $4\times$. The best result for each image in the data set and each metric is highlighted in bold.}
\begin{tabular}{|l|c|c|c|c|c|c|}
\hline
\bf Image 		& \multicolumn{2}{|c|}{\bf Bicubic interpolation} 	& \multicolumn{2}{|c|}{\bf Tang et al.~\cite{Tang-IJMLC-2011}} & \multicolumn{2}{|c|}{\bf CocoNet (10 layers)}\\
\cline{2-7}
                & \bf \;\;\;PSNR\;\;\;  & \bf SSIM              & \bf \;\;\;PSNR\;\;\;  & \bf \;\;\;SSIM\;\;\;  & \bf \;\;\;PSNR\;\;\;      & \bf SSIM\\
\hline
baby 			& $29.39$		        & $\mathbf{0.8162}$     & $\mathbf{29.70}$      & -         & $26.88$					& $0.7146$\\
bird            & $26.84$		        & $\mathbf{0.8265}$     & $\mathbf{27.84}$      & -         & $26.71$					& $0.7816$\\
butterfly       & $19.93$				& $0.6779$              & $20.61$               & -         & $\mathbf{20.92}$			& $\mathbf{0.7229}$\\
head            & $28.48$				& $0.6786$              & $\mathbf{29.83}$		& -	        & $29.26$                   & $\mathbf{0.7016}$\\
woman           & $24.17$		        & $0.7979$              & $24.46$               & -         & $\mathbf{25.33}$			& $\mathbf{0.8150}$\\
\hline
\end{tabular}
\end{center}
}
\vspace{-0,6cm}
\end{table}

\begin{figure}[!t]
\begin{center}
\includegraphics[width=0.97\columnwidth]{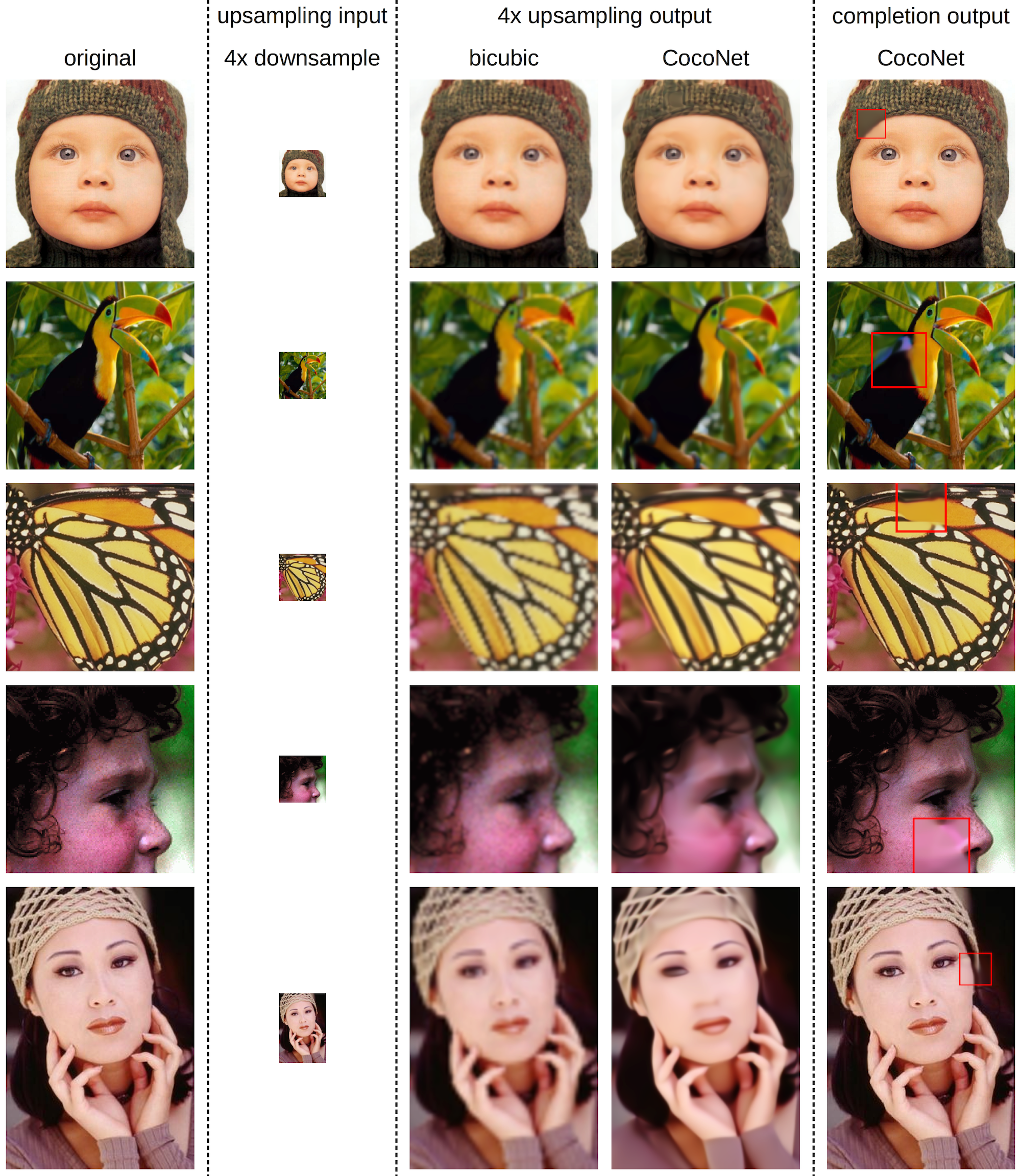}
\end{center}
\vspace*{-0.45cm}
\caption{Image upsampling and completion results on Set5~\cite{Bevilacqua-BMVC-2012}. For the image upsampling experiments, the original images are first downsampled by a scale factor of $4\times$. For the image completion experiments, a random patch (surrounded by a red bounding box) from the original image is removed during training. For both applications, the CocoNet architecture has $15$ layers. Best viewed in color.}
\label{fig_upsampling_completion}
\vspace*{-0.5cm}
\end{figure}

For image resampling, we consider a scale factor of $4\times$, a standard choice for image super-resolution approaches~\cite{Bevilacqua-BMVC-2012,Kim-CVPR-2016,Ledig-CVPR-2016,Tang-IJMLC-2011,Yamanaka-ICONIP-2017}. Although we use the same scale factor, it is not fair to compare our approach with image super-resolution approaches~\cite{Kim-CVPR-2016,Ledig-CVPR-2016,Sajjadi-ICCV-2017,Tang-IJMLC-2011,Yamanaka-ICONIP-2017} that are trained on both low-resolution (downsampled) and high-resolution (upsampled) patches sampled from the input image or from additional training images, since we train our method solely on the downsampled image. Nevertheless, we compare our CocoNet architecture, that consists of $10$ layers each of $200$ neurons, with a single-image super-resolution approach~\cite{Tang-IJMLC-2011} and bicubic interpolation. The corresponding results are presented in Table~\ref{Tab_Resampling_Results}. In terms of PSNR, we obtain better results for two images (\emph{butterfly} and \emph{woman}). In terms of SSIM, we obtain better results than bicubic interpolation for three images. The upsampling results of bicubic interpolation and CocoNet are depicted in Figure~\ref{fig_upsampling_completion}. In the \emph{bird}, the \emph{butterfly} and the \emph{woman} images, we notice that the lines produced by CocoNet are smoother than those resulted after bicubic interpolation.

\vspace{-0,2cm}
\subsection{Image Completion Results}
\vspace{-0,1cm}

We present image completion results on the last column of Figure~\ref{fig_upsampling_completion}. The results are obtained with a CocoNet architecture of $10$ layers, each of $200$ neurons. For each image in Set5, we select a random patch that is removed at training time. Consequently, CocoNet has to automatically complete the respective patch at test time. The qualitative results depicted in Figure~\ref{fig_upsampling_completion} indicate that CocoNet is generally able to complete the high-level shapes (lines and curves) in the missing regions. However, without any additional training sources, CocoNet is not able to generate (or reproduce) the texture details of the missing regions.

\vspace{-0,2cm}
\section{Conclusion and Further Work}
\label{sec_Conclusion}
\vspace{-0,3cm}

In this paper, we proposed CocoNet, a neural approach that learns to map the pixel coordinates (in a single input image) to the corresponding RGB values. We conducted experiments on various low-level image processing tasks, namely image denoising, image upsampling (super-resolution) and image completion, in order to demonstrate the usability of our approach. In future work, we aim to address other possible applications of our neural network, e.g. image compression. We also aim to try out a convolutional architecture instead of a fully-connected one (the preliminary results in this direction are promising).

\bibliography{references}{} 
\bibliographystyle{splncs03}

\end{document}